\documentclass[10pt,twocolumn,letterpaper]{article}

\usepackage{authblk}

\usepackage[pagenumbers]{wacv} 

\usepackage{graphicx}
\usepackage{amsmath}
\usepackage{amssymb}
\usepackage{booktabs}

\usepackage[pagebackref,breaklinks,colorlinks]{hyperref}

\usepackage[utf8]{inputenc} %
\usepackage[T1]{fontenc}    %
\usepackage{hyperref}       %
\usepackage{url}            %
\usepackage{booktabs}       %
\usepackage{amsfonts}       %
\usepackage{nicefrac}       %
\usepackage{microtype}      %
\usepackage{xcolor}         %
\usepackage{float}
\usepackage{multirow}
\usepackage[font={small}]{caption}
\usepackage[font={small}]{subcaption}

\usepackage{graphicx}
\usepackage{pifont}

\usepackage{arydshln}

\definecolor{cssgreen}{rgb}{0.0, 0.5, 0.0}
\definecolor{cssred}{rgb}{1, 0, 0.0}
\definecolor{DarkGreen}{rgb}{0.43, 0.68, 0.28}

\newcommand{\dataset}{COCO-FP}

\newcommand{\cmark}{\ding{51}} %

\usepackage[capitalize]{cleveref}
\crefname{section}{Sec.}{Secs.}
\Crefname{section}{Section}{Sections}
\Crefname{table}{Table}{Tables}
\crefname{table}{Tab.}{Tabs.}

\def\wacvPaperID{2369} %
\def\confName{WACV}
\def\confYear{2025}

\begin{document}

\title{From COCO to COCO-FP: A Deep Dive into Background False Positives for COCO Detectors}

\author{Longfei Liu\textsuperscript{1,2} \quad Wen Guo\textsuperscript{3} \quad Shihua Huang\textsuperscript{1} \quad Cheng Li\textsuperscript{1} \quad Xi Shen\textsuperscript{1$\dag$} \quad \\
\textsuperscript{1}Intellindust, China \\
\textsuperscript{2}Guilin University Of Electronic Technology \\
\textsuperscript{3}ETH Zurich \\

}

\maketitle

\renewcommand{\thefootnote}{\fnsymbol{footnote}}

\footnotetext[2]{Corresponding Author.}
\renewcommand{\thefootnote}{\arabic{footnote}}
\begin{abstract}

Reducing false positives is essential for enhancing object detector performance, as reflected in the mean Average Precision (mAP) metric. Although object detectors have achieved notable improvements and high mAP scores on the COCO dataset, analysis reveals limited progress in addressing false positives caused by non-target visual clutter — background objects not included in the annotated categories. This issue is particularly critical in real-world applications, such as fire and smoke detection, where minimizing false alarms is crucial. In this study, we introduce \dataset, a new evaluation dataset derived from the ImageNet-1K dataset, designed to address this issue. By extending the original COCO validation dataset, \dataset~specifically assesses object detectors' performance in mitigating background false positives. Our evaluation of both standard and advanced object detectors shows a significant number of false positives in both closed-set and open-set scenarios. For example, the AP50 metric for YOLOv9-E decreases from 72.8 to 65.7 when shifting from COCO to \dataset. The dataset is available at \url{https://github.com/COCO-FP/COCO-FP}.

\end{abstract}

\section{Introduction}
\label{sec:intro}

\begin{figure}[!t]
  \centering
  \includegraphics[width=1\linewidth]{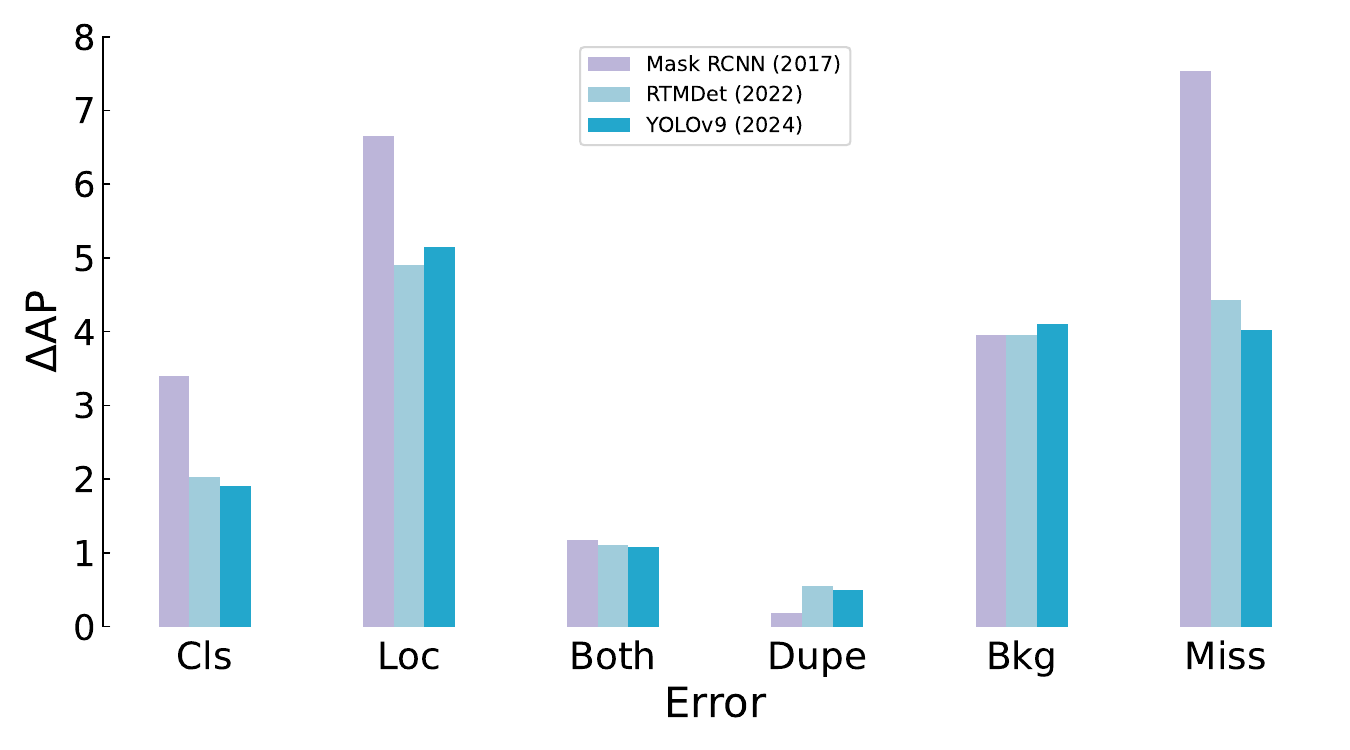}
  \caption{\textbf{Summary of errors for object detectors on the COCO~\cite{lin2014microsoft} dataset.} $\Delta\text{AP}$ illustrates the absolute contribution of each error type, computed from TIDE~\cite{tide-eccv2020}. We apply error analysis to Mask RCNN~\cite{He_2017_ICCV}, RTMDet~\cite{lyu2022rtmdet}, and YOLOv9~\cite{wang2024yolov9}, with their mAP scores reaching 40.1, 52.8, and 55.6, respectively, on the COCO Val dataset.}
  \label{fig:error}
\end{figure}

\begin{figure}[!t]
  \centering
    \begin{subfigure}[b]{\linewidth}
         \centering
         \includegraphics[width=0.9\linewidth]{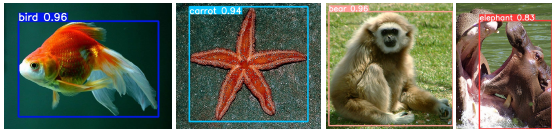}
         \caption{False positive predictions of YOLOv9-E~\cite{wang2024yolov9} on the proposed \dataset~dataset.}
         \label{fig:teaser_a}
         \centering
        \includegraphics[width=0.9\linewidth]{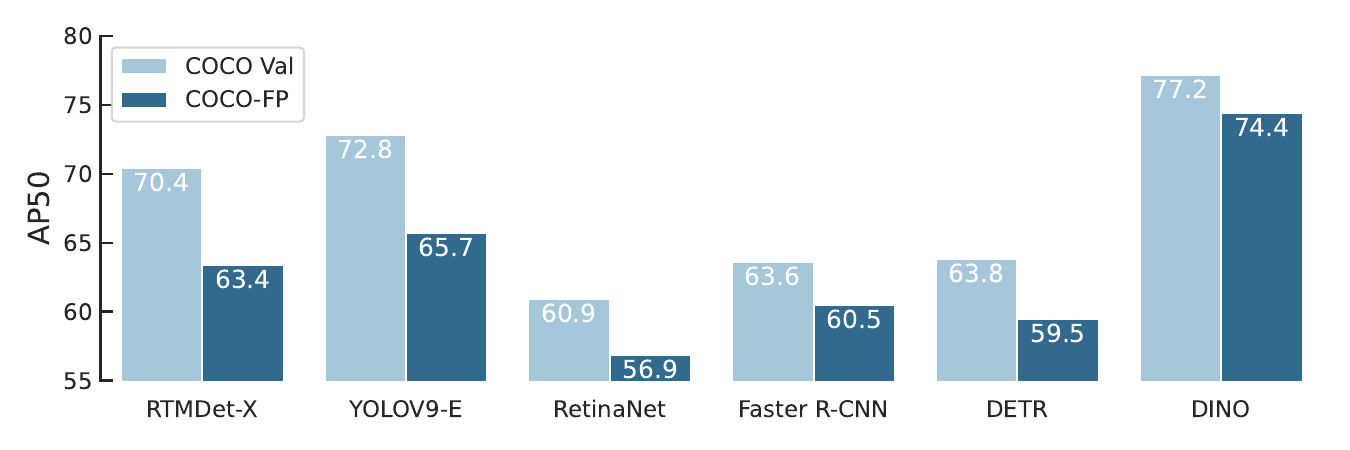}
         \caption{AP50 for closed-set object detectors on COCO~\cite{lin2014microsoft} and \dataset.}
         \label{fig:teaser_b}
         \centering
        \includegraphics[width=0.9\linewidth]{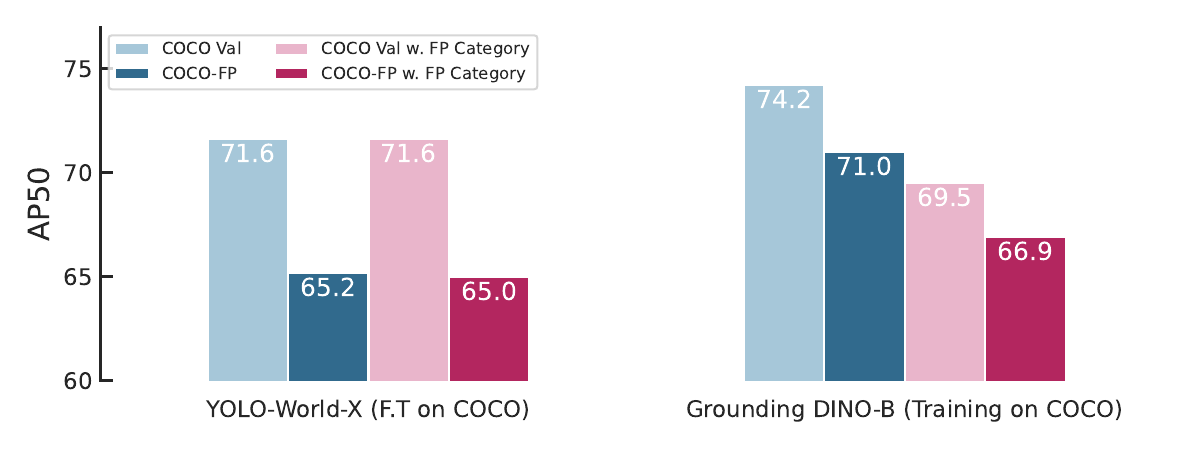}
         \caption{AP50 for open-set object detectors on COCO~\cite{lin2014microsoft} and \dataset.} 
         \label{fig:teaser_c}
    \end{subfigure}
    \begin{subfigure}[b]{\linewidth}

         \label{fig:teaser}
     \end{subfigure}
  \caption{\textbf{Qualitative and quantitative analysis of false positives for object detectors.} \textit{(a)} False positive examples of  YOLOv9~\cite{wang2024yolov9} on our \dataset~dataset. Note that the confidence scores of the detected bounding boxes are important. \textit{(b)} The performance of closed-set detectors. It's noteworthy that transitioning to \dataset~results in a significant decrease in performance. \textit{(c)} The performance of open-set detectors. The term "w. FP category" refers to leveraging the names of extra image categories as an additional text prompt for detection purposes. YOLO-World-X~\cite{Cheng2024YOLOWorld} undergoes pre-training on a large-scale dataset followed by fine-tuning on the COCO~\cite{lin2014microsoft} dataset. While Grounding DINO-B~\cite{liu2023grounding} is trained on a large-scale dataset including COCO~\cite{lin2014microsoft}.}
  \label{fig:analysis }
\end{figure}

Object detection is a fundamental task in the field of computer vision, aimed at accurately locating and classifying objects within images. In recent years, the field has witnessed remarkable progress, with significant advancement in mean Average Precision (mAP), a key metric for assessing object detection performance. This improvement leading to numerous real-world applications~\cite{li2015convolutional,nitsch2021out,zhu2006fast,wu2017squeezedet}. The requirements for object detectors vary across different scenarios, especially in task-specific environments where detectors must focus solely on relevant objects without being affected by background.

However, object detectors still encounter challenges in practical applications. A particularly persistent issue is the prevalence of false positive, which can significantly undermine the reliability and efficacy of detection systems. The causes of false positives are multifaceted, and relying solely on mAP does not provide a clear assessment of the specific impact of each error type. TIDE~\cite{tide-eccv2020} offers a detailed breakdown of mAP by categorizing errors into six types: classification, localization, combined classification and localization, duplicate detection, background, and missed Ground Truth errors. Of these, classification and localization errors encompass both false positives and false negatives, while background errors refer to false positives caused by non-target visual clutter (objects outside the annotated categories) and missed errors to false negatives. As depicted in Figure~\ref{fig:error}, classification, localization, missed, and background errors are the most prominent. Notably, while other error types decrease as detector performance improves, background errors remain consistently high, underscoring the persistent vulnerability of object detection models to false positives despite overall advancements.

To reveal this problem concretely, instead of testing the COCO-trained detectors on a complicated dataset, we test them on the ImageNet-1K dataset~\cite{deng2009imagenet}. ImageNet samples are generally considered relatively simple, as most images capture a significant main object with a clear background. we use the YOLOv9-E~\cite{wang2024yolov9}, a state-of-the-art detector, to predict some images from the ImageNet dataset. Note that none of the images contain objects from COCO pre-defined classes, however, the YOLOv9-E~\cite{wang2024yolov9} still produces bounding box predictions with high confidence scores. In many real-world applications, detectors are designed to focus on specific categories, such as person detection~\cite{zhu2006fast}, face detection~\cite{li2015convolutional}, and fire detection~\cite{wu2019intelligent} etc. In such cases, the objective is to accurately identify objects from predefined classes, as false positive predictions can cause serious issues for downstream tasks. For example, an AI-based fire suppression system links the activation of fire extinguishers to fire detection outputs. False positive predictions in this context would result in unnecessary costs for the entire system.

To better reveal this issue and benchmark the performance of object detectors in the context of background false positives, we collected a high-quality dataset leveraging images from the ImageNet-1K~\cite{deng2009imagenet}. We aimed to select images devoid of any objects defined in COCO, thereby potentially inducing incorrect predictions in COCO-trained detectors. We employ one of the state-of-the-art detectors Co-DETR~\cite{zong2022detrs} trained on COCO (achieving 64.1 AP on COCO validation set) to get predictions for ImageNet and meticulously conduct the entire data collection process. This involved carefully removing semantic-overlap categories (exact duplicates in COCO or semantically related to COCO categories), eliminating irrelevant categories (which produce very few bounding boxes predictions), applying manual image-level filtering to ensure the dataset to be accurate, and finally manually refining the dataset to ensure both high diversity and balanced distribution. This rigorous pipeline yields 3,772 images spanning 50 different categories (Figure~\ref{fig:pipeline}). In conjunction with the original COCO validation set, we introduce a novel evaluation dataset, denoted as \dataset.

Despite selecting samples based on predictions from Co-DETR~\cite{zong2022detrs}, other object detectors still produce a significant number of false positive (FP) predictions on our \dataset dataset. For benchmarking purposes, we evaluated both traditional deep learning-based and state-of-the-art closed-set object detectors on our proposed dataset. Most detectors exhibited a substantial number of FPs, with performance notably declining when transitioning from COCO to \dataset~(Figure~\ref{fig:teaser_b}). We also assessed recent open-set object detectors, such as YOLO-World~\cite{Cheng2024YOLOWorld}, GLIP~\cite{li2022grounded}, and Grounding DINO~\cite{liu2023grounding} (Figure~\ref{fig:teaser_c}). These detectors, which identify objects based on provided categories, also experienced significant performance drops with \dataset. Additionally, we evaluated these open-set detectors using text prompts based on the FP categories. While YOLO-World-X~\cite{Cheng2024YOLOWorld} maintained similar performance, Grounding DINO-B~\cite{liu2023grounding} exhibited a significant decline. This highlights that current open-set detection methods, designed for open-world object detection, struggle to generalize to the categories within our \dataset~dataset.

To conclude, our contributions are as follows: %
\begin{itemize}
    \item We introduce \dataset, a complementary evaluation dataset to COCO~\cite{lin2014microsoft}, derived from ImageNet-1K~\cite{deng2009imagenet}. \dataset~is specifically designed to evaluate object detectors' ability to mitigate false positives caused by non-target visual clutter. This dataset challenges both traditional closed-set and modern open-set detectors, highlighting significant performance declines compared to COCO.
    \item We conduct an extensive evaluation of both traditional deep learning-based and state-of-the-art object detectors, including open-set models. Our findings demonstrate that, despite advancements, current detection methods, including those designed for open-world scenarios, struggle to generalize effectively to the challenging categories presented in \dataset.
\end{itemize}

Our dataset is available at \url{https://github.com/COCO-FP/COCO-FP}.

\section{Related Work}
\label{sec:related_work}

\paragraph{Datasets for object detection.} In object detection, several benchmark datasets play a critical role in evaluating algorithm performance. The PASCAL VOC dataset~\cite{everingham2010pascal}, with around 11K images across 20 object categories, laid the foundation for benchmark evaluation. The COCO dataset~\cite{lin2014microsoft} expanded this by including 80 object categories and over 120K images, including 5K validation images, with a focus on detecting objects in diverse and cluttered scenes that closely mimic real-world conditions. This has made COCO a crucial benchmark in computer vision. The Objects365-V2 dataset~\cite{shao2019objects365} further raised the challenge with 365 object categories, over 2 million images, and more than 30 million labeled bounding boxes. The Open Images V7 dataset~\cite{kuznetsova2020open} scales this up even further, containing approximately 9 million images with 16 million bounding boxes across 600 categories. The LVIS dataset~\cite{gupta2019lvis} introduces a long-tail distribution with 164K images and around 2 million instances for 1,203 categories.

Among these, COCO remains the most widely used benchmark due to its balance of scale and computational requirements. Many works have extended COCO to explore new challenges. For example, COCO-Stuff~\cite{caesar2018coco} adds pixel-level annotations for 91 object and background categories, COCONut~\cite{deng2024coconut} expands it to 383K images for large-scale segmentation tasks, and COCO-C~\cite{michaelis2019benchmarking} introduces synthetic corruptions like noise and blur to evaluate robustness. COCO-O~\cite{mao2023coco} introduces domain shifts with approximately 7K images from such as paintings, cartoons, and sketches to test model robustness under natural distribution changes.

In contrast to these extensions, our proposed dataset offers a unique challenge by introducing additional images with no overlap in object categories with COCO. This enables a more rigorous evaluation of object detectors' ability to handle false positives from non-target visual clutter, a critical issue for real-world applications.

\paragraph{Closed-set object detection.} 
Closed-set object detection focuses on detecting objects from predefined categories known during training. Recently, significant advancements have been made in this field. Early influential methods include anchor-based frameworks like Fast R-CNN~\cite{girshick2015fast}, which introduced end-to-end detection with ROI pooling, and Faster R-CNN~\cite{ren2015faster}, which incorporated Region Proposal Networks (RPN) for efficient bounding box prediction. Anchor-free frameworks have since gained traction, such as CenterNet~\cite{duan2019centernet}, FCOS~\cite{tian2019fcos}, and the YOLO series~\cite{redmon2016you,redmon2017yolo9000,redmon2018yolov3,ge2021yolox,wang2023yolov7,wang2024yolov9}, as well as SSD~\cite{liu2016ssd}. These methods predict bounding boxes and class probabilities directly from full images in one pass, resulting in faster and more efficient detection. For example, YOLOX~\cite{ge2021yolox} enhances accuracy and inference speed with a decoupled detection head and adaptive strategies. CenterNet~\cite{duan2019centernet} identifies objects by their center points, while FCOS~\cite{tian2019fcos} employs a pixel-wise approach on feature maps. The introduction of transformer architectures~\cite{vaswani2017attention,dosovitskiy2020image,carion2020end,zhu2020deformable} has further advanced object detection. DETR~\cite{carion2020end} reformulated object detection as a set prediction problem using transformers for end-to-end detection, and Deformable DETR~\cite{zhu2020deformable} improved it with deformable attention mechanisms~\cite{dai2017deformable} for better object positioning. The recent Co-DETR~\cite{zong2022detrs} achieves state-of-the-art performance by introducing a collaborative hybrid assignments training scheme within the DETR framework. These advancements have significantly enhanced detection accuracy and speed, broadening the applicability of object detectors. %

\paragraph{Open-set object detection}

Open-set Object detection is an emerging research area in computer vision that addresses the limitations of traditional object detectors when encountering unknown classes during inference. OV-DETR~\cite{zang2022open} proposes a detector based on DETR~\cite{carion2020end} and uses pre-trained visual-language model~\cite{radford2021learning} generation queries to condition the transformer decoder. GLIP~\cite{li2022grounded} reformulate object detection based on phrase grounding and align phrase and object-level visual representations through extensive pretraining on large-scale datasets.This enables the GLIP~\cite{li2022grounded} model to perform well on downstream tasks. Grounding DINO~\cite{liu2023grounding} combines grounded pre-training with DINO~\cite{zhang2022dino}  by performing vision-language modality fusion.YOLO-World~\cite{Cheng2024YOLOWorld} employs YOLOv8~\cite{Jocher_Ultralytics_YOLO_2023} as its backbone and enhances YOLO with open-set detection capability by utilizing cross-modality fusion for both image and text features. It demonstrates real-time inference capabilities that other open-set methods fail to achieve, while maintaining comparable performance. GenerateU~\cite{lin2024generateu} proposes a new setting known as generative open-ended object detection, which addresses the issue of needing predefined categories during inference by reframing object detection as a generative task.

\section{Our Dataset:~\dataset}
\label{sec:dataset}
\begin{figure*}[!t]
  \centering
  \includegraphics[width=1\linewidth]{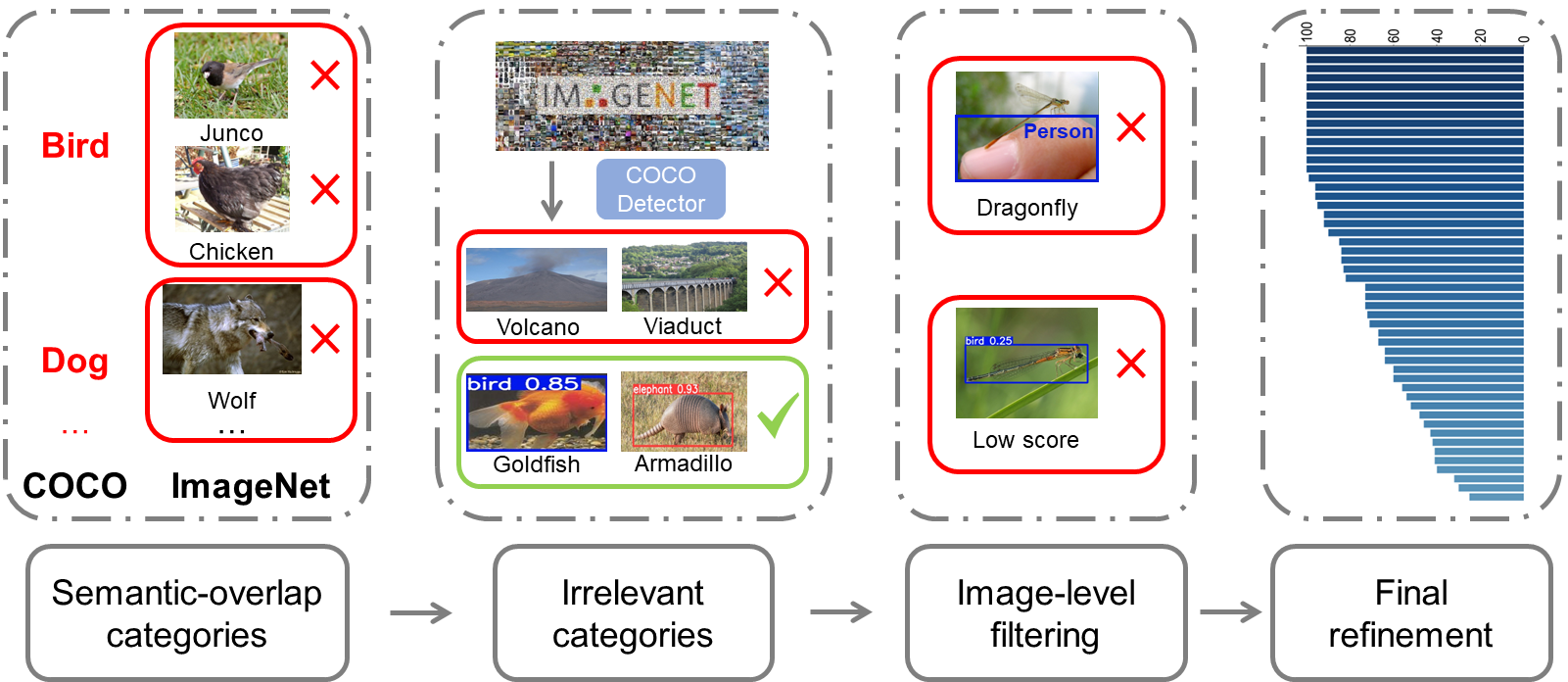}
  \caption{\textbf{Dataset collection pipeline.} Initially, categories in ImageNet~\cite{deng2009imagenet} that semantically overlap with those in COCO~\cite{lin2014microsoft} are excluded. Subsequently, the COCO-trained detector Co-DETR~\cite{zong2022detrs} is utilized to identify categories that produce false positive predictions. Next, all images of the remaining categories are processed through Co-DETR~\cite{zong2022detrs}, with each image being filtered individually. Finally, to ensure dataset diversity and balanced distribution, we retain at most 100 images per category and only retain certain categories in case multiple categories in ImageNet~\cite{deng2009imagenet} are misidentified as the same category in COCO~\cite{lin2014microsoft}.}
  \label{fig:pipeline}
\end{figure*}

Despite the significant improvements in overall performance of object detectors, progress in reducing background false positives~\cite{tide-eccv2020}, particularly those caused by non-target visual clutter (as shown in Figure~\ref{fig:error}), remains limited. To address this issue, we introduce an updated version of the COCO validation dataset~\cite{lin2014microsoft}, designed to enable a more precise evaluation of object detector performance in real-world scenarios. In this section, we present our dataset, \dataset, created to tackle the challenge of background-induced false positives. This dataset aims to support the development of methods that enhance the robustness and reliability of detection models by effectively mitigating background false positives.

\paragraph{Building \dataset~from ImageNet~\cite{deng2009imagenet}}

We construct \dataset~by extending the official COCO validation dataset~\cite{lin2014microsoft} with images from the large-scale ImageNet dataset~\cite{deng2009imagenet}. Our goal is to select images that do not contain any COCO-defined objects but could potentially lead to false positive predictions in object detectors. To achieve this, we develop a pipeline to process the ImageNet-1K dataset, as illustrated in Figure~\ref{fig:pipeline}. The pipeline consists of the following four steps:

\begin{itemize}
 \item Remove semantic-overlap categories: We start by excluding categories that semantically overlap with categories defined in COCO~\cite{lin2014microsoft}. This involves three scenarios: \textit{i)} categories that are exact duplicates in COCO, such as "Banana" and "Broccoli"; \textit{ii)} categories that are subcategories in COCO, where broader concepts like "Bird" encompass specific types like "Chicken", "Duck," and "Goose"; and \textit{iii)} categories that share similar traits and may cause disagreement, like "Compass" in ImageNet~\cite{deng2009imagenet} and "Clock" in COCO. After this process, approximately 600 categories remain.

\item Discard irrelevant categories: We eliminate those that significantly deviate from COCO categories and are less likely to cause errors for object detectors trained on COCO~\cite{lin2014microsoft}. To achieve this, we utilize the latest state-of-the-art object detector Co-DETR~\cite{zong2022detrs} with Swin-L backbone~\cite{liu2021swin}, which manages to achieve 64.1 AP on the COCO validation dataset. By randomly sampling 50 images per category and processing them with Co-DETR~\cite{zong2022detrs}, we remove categories with no or very few bounding box predictions.

\item Image-level filtering: In this step, we manually check the collected images. Specifically, we first process all remaining images with Co-DETR~\cite{zong2022detrs} and retain only those containing bounding box with a confidence score greater than 0.3. Here we choose a relatively high score to avoid getting too many false positive predictions.
Then, we manually check every collected image and remove the images containing any COCO objects or multiple objects (Figure~\ref{fig:pipeline}).

\item Final refinement: To ensure dataset diversity, we retain at most the first 100 images with the highest confidence scores in each category. Additionally, when multiple categories in ImageNet~\cite{deng2009imagenet} are misidentified as the same category in COCO~\cite{lin2014microsoft}, we selectively retain only the categories with the most false positive prediction. For example, koalas, sea lions, and dugongs are often misidentified as bears. In such cases, we choose to retain only the images of dugongs, as they contain the most false positive samples compared to the other categories.

\end{itemize}

This pipeline gives us in total 3,772 images from 50 categories. The number of images in each category is detailed in the Appendix~\ref{sec:distribution}. The dataset is high-quality and contains comparable number of images to the COCO 5K-image validation set, making it suitable for efficient and effective evaluation with most object detectors.

Interestingly, although our dataset was selected using a single object detector, as shown in Section~\ref{sec:benchmark}, the Co-DETR~\cite{zong2022detrs}, other object detectors also perform poorly on it, including both anchor-based and anchor-free models using different architectures (CNN or Transformer), indicating that this issue is common to all object detectors. Visual examples of images detected by different detectors are shown in Figure~\ref{fig:visual}, with additional visualizations provided in the Appendix~\ref{sec:visualizations}.

After combining with COCO 5K-image validation set, we propose \dataset, which consists of 8,772 images. Unlike standard adversarial samples that generate perturbations to induce incorrect predictions~\cite{wu2020making, song2018physical, xu2020adversarial}, our dataset contains only natural images, thereby reflecting performance in real-world applications.

\begin{table*}[t]
    \begin{center}
		\resizebox{2\columnwidth}{!}{
			\begin{tabular}{l|l|l|c|c||c|c|c||c|c}
				\hline 
				\multirow{ 2}{*}{Detector}  & \multirow{ 2}{*}{Backbone} & \multirow{ 2}{*}{Neck / Decoder}& Image & \multirow{ 2}{*}{Epochs} & \multirow{ 2}{*}{Params} & \multicolumn{2}{c||}{COCO Val~\cite{lin2014microsoft}} & \multicolumn{2}{c}{\dataset} \\ 
                 & & & Resolution  & & & AP  & AP50 & AP  & AP50 \\ 
    
                \hline
				\multicolumn{10}{c}{\textbf{CNN-based and Anchor-free Approaches}}           \\
                FCOS~\cite{tian2019fcos}   & ResNeXt-64x4d-101~\cite{xie2017aggregated}  & FPN~\cite{lin2017feature} & 1333$\times$800  &24  & 90.2M &  42.8&  62.6  & 39.5 \bf \textcolor{cssred}{$\downarrow$ 3.3}      & 57.3 \bf \textcolor{cssred}{$\downarrow$ 5.3} \\
                CenterNet~\cite{duan2019centernet}  & ResNet-50~\cite{he2016deep}  & FPN~\cite{lin2017feature} & 1333$\times$800 & 12 & 32.3M &  40.2  &  58.3 & 37.3 \bf \textcolor{cssred}{$\downarrow$ 2.9} & 53.7 \bf \textcolor{cssred}{$\downarrow$ 4.6} \\
                
                YOLOX-S~\cite{ge2021yolox}    & CSPDarkNet~\cite{bochkovskiy2020yolov4} &  PAFPN~\cite{liu2018path} & 640$\times$640 &300 & 8.9M &  40.5 &  59.5 & 37.2 \bf \textcolor{cssred}{$\downarrow$ 3.3}  & 54.5 \bf \textcolor{cssred}{$\downarrow$ 5.0}  \\
                RTMDet-S~\cite{lyu2022rtmdet}    &  CSPNeXt~\cite{wang2020cspnet} & PAFPN~\cite{liu2018path} & 640$\times$640  &300 & 8.9M  &  44.6  &  61.8 & 41.0 \bf \textcolor{cssred}{$\downarrow$ 3.6}        & 56.4 \bf \textcolor{cssred}{$\downarrow$ 5.4}  \\
                YOLOv8-S~\cite{Jocher_Ultralytics_YOLO_2023}  &  CSPDarkNet~\cite{bochkovskiy2020yolov4} & PAFPN~\cite{liu2018path} & 640$\times$640  & 500 & 11.2M  & 44.9   &  61.8 & 40.9 \bf \textcolor{cssred}{$\downarrow$ 4.0}  & 56.0 \bf \textcolor{cssred}{$\downarrow$ 5.8}\\
                YOLOv9-S~\cite{wang2024yolov9}  &  CSPDarkNet~\cite{bochkovskiy2020yolov4} & PAFPN~\cite{liu2018path} & 640$\times$640  & 500 & 7.1M  & 46.8   &  63.4 & 41.9 \bf\textcolor{cssred}{$\downarrow$ 4.9}        & 56.6 \bf\textcolor{cssred}{$\downarrow$ 6.8}  \\
                YOLOv10-S~\cite{wang2024yolov10} & CSPDarkNet~\cite{bochkovskiy2020yolov4} & PAFPN~\cite{liu2018path} & 640$\times$640 & 500 & 7.2M&46.2&63.0&42.0 \bf\textcolor{cssred}{$\downarrow$ 4.2}&57.2 \bf\textcolor{cssred}{$\downarrow$ 5.8}\\

                YOLOX-X~\cite{ge2021yolox}    & CSPDarkNet~\cite{bochkovskiy2020yolov4} &  PAFPN~\cite{liu2018path} & 640$\times$640 &300 & 99.1M   &  50.9 &  69.0 & 45.4 \bf \textcolor{cssred}{$\downarrow$ 5.5}  & 61.6 \bf \textcolor{cssred}{$\downarrow$ 7.4}  \\
                RTMDet-X~\cite{lyu2022rtmdet}    &  CSPNeXt~\cite{wang2020cspnet} & PAFPN~\cite{liu2018path} & 640$\times$640  &300 & 94.9M  &  52.8  &  70.4 & 47.8 \bf \textcolor{cssred}{$\downarrow$ 5.0}        & 63.4 \bf \textcolor{cssred}{$\downarrow$ 7.0}  \\
                YOLOv8-X~\cite{Jocher_Ultralytics_YOLO_2023}  &  CSPDarkNet~\cite{bochkovskiy2020yolov4} & PAFPN~\cite{liu2018path} & 640$\times$640  & 500 & 68.2M  & 54.0   &  71.0 & 49.2 \bf \textcolor{cssred}{$\downarrow$ 4.8}        & 64.5 \bf \textcolor{cssred}{$\downarrow$ 6.5}  \\
                YOLOv9-E~\cite{wang2024yolov9}  &  CSPDarkNet~\cite{bochkovskiy2020yolov4} & PAFPN~\cite{liu2018path} & 640$\times$640  & 500 & 57.3M  & 55.6   &  72.8 & 50.2 \bf\textcolor{cssred}{$\downarrow$ 5.4}        & 65.7 \bf\textcolor{cssred}{$\downarrow$ 7.1}  \\
                YOLOv10-X~\cite{wang2024yolov10} & CSPDarkNet~\cite{bochkovskiy2020yolov4} & PAFPN~\cite{liu2018path} & 640$\times$640 & 500 & 29.5M&54.4&71.3&49.4 \bf\textcolor{cssred}{$\downarrow$ 5.0}&64.7 \bf\textcolor{cssred}{$\downarrow$ 6.6}\\
                
                \hline
				\multicolumn{10}{c}{\textbf{CNN-based and Anchor-based Approaches}} \\
                Faster R-CNN~\cite{ren2015faster} & ResNeXt-64x4d-101~\cite{xie2017aggregated}  & FPN~\cite{lin2017feature} & 1333$\times$800 &36& 99.6M &   43.2 & 63.6   & 41.2 \bf \textcolor{cssred}{$\downarrow$ 2.0}   & 60.5 \bf \textcolor{cssred}{$\downarrow$ 3.1} \\
                Cascade R-CNN~\cite{cai2018cascade}   & ResNeXt-64x4d-101~\cite{xie2017aggregated}  & FPN~\cite{lin2017feature} & 1333$\times$800 & 12 & 127.3M & 44.7  &  63.6 &42.9  \bf \textcolor{cssred}{$\downarrow$ 1.8}    &60.7 \bf \textcolor{cssred}{$\downarrow$ 2.9}  \\
                RetinaNet~\cite{lin2017focal}   & ResNeXt-64x4d-101~\cite{xie2017aggregated}  & FPN~\cite{lin2017feature} & 1333$\times$800 & 36& 95.6M & 41.0  &  60.9 & 38.6  \bf \textcolor{cssred}{$\downarrow$ 2.4}    &56.9 \bf \textcolor{cssred}{$\downarrow$ 4.0}  \\
                
                YOLOv5-S~\cite{yolov5}   & CSPDarkNet~\cite{bochkovskiy2020yolov4}  & PAFPN~\cite{liu2018path} & 640$\times$640 & 300& 7.2M & 37.4  &  56.8 & 35.4 \bf \textcolor{cssred}{$\downarrow$ 2.0}    &53.5 \bf \textcolor{cssred}{$\downarrow$ 3.3}  \\
                YOLOv5-M~\cite{yolov5}   & CSPDarkNet~\cite{bochkovskiy2020yolov4}  & PAFPN~\cite{liu2018path} & 640$\times$640 & 300& 21.2M & 45.3  &  64.1 & 42.9 \bf \textcolor{cssred}{$\downarrow$ 2.4}    &60.6 \bf \textcolor{cssred}{$\downarrow$ 3.5}  \\
                YOLOv5-L~\cite{yolov5}   & CSPDarkNet~\cite{bochkovskiy2020yolov4}  & PAFPN~\cite{liu2018path} & 640$\times$640 & 300& 46.5M & 49.0  &  67.4 & 46.2 \bf \textcolor{cssred}{$\downarrow$ 2.8}    &63.3 \bf \textcolor{cssred}{$\downarrow$ 4.1}  \\
                YOLOv5-X~\cite{yolov5}   & CSPDarkNet~\cite{bochkovskiy2020yolov4}  & PAFPN~\cite{liu2018path} & 640$\times$640 & 300& 86.7M & 50.7  &  68.9 & 47.7 \bf \textcolor{cssred}{$\downarrow$ 3.0}    &64.2 \bf \textcolor{cssred}{$\downarrow$ 4.7}  \\
                YOLOv7-X~\cite{wang2023yolov7}   & CSPDarkNet~\cite{bochkovskiy2020yolov4}  & PAFPN~\cite{liu2018path} & 640$\times$640 & 300 & 71.3M & 52.9  &  71.1 & 49.0 \bf \textcolor{cssred}{$\downarrow$ 3.9}    &65.6 \bf \textcolor{cssred}{$\downarrow$ 5.5}  \\
                 \hline
				\multicolumn{10}{c}{\textbf{Transofrmer-based Approaches}} \\
                DETR~\cite{carion2020end}  & ResNet-101~\cite{he2016deep}  & Decoder & 1333$\times$800 & 500 & 60.6M  &  43.5  &  63.8 & 40.6 \bf \textcolor{cssred}{$\downarrow$ 2.9}         & 59.5 \bf \textcolor{cssred}{$\downarrow$ 4.3} \\
                Deformable-DETR~\cite{zhu2020deformable}  & ResNet-50~\cite{he2016deep}  & Decoder & 1333$\times$800 & 50 & 41.2M  &  47.0  &  66.1 & 43.7 \bf\textcolor{cssred}{$\downarrow$ 3.3}         & 61.5 \bf\textcolor{cssred}{$\downarrow$ 4.6} \\
                DAB-DETR~\cite{liu2022dabdetr}  & ResNet-50~\cite{he2016deep}  & Decoder & 1333$\times$800 &50 & 43.7M  &  42.3  &  62.9 & 38.7 \bf\textcolor{cssred}{$\downarrow$ 3.6}         & 57.6 
 \bf\textcolor{cssred}{$\downarrow$ 5.3}\\
                DINO~\cite{zhang2022dino}  & Swin-L~\cite{liu2021swin}  & Decoder & 1333$\times$800 &36 & 218.3M  &  58.4  &  77.2 & 56.3 \bf \textcolor{cssred}{$\downarrow$ 2.1}        & 74.4 \bf\textcolor{cssred}{$\downarrow$ 2.8} \\
                RT-DETR~\cite{lv2023detrs}  & ResNet-101~\cite{he2016deep}  & Decoder & 640$\times$640 &72 & 76.0M  &  54.3  &  72.8 & 48.6 \bf \textcolor{cssred}{$\downarrow$ 5.7}  & 65.0 \bf \textcolor{cssred}{$\downarrow$ 7.8} \\
                Co-DETR~\cite{zong2022detrs}  & Swin-L~\cite{liu2021swin}  & Decoder & 2048$\times$1280 &16 & 235.4M  &  64.1  &  81.3 & 60.3 \bf \textcolor{cssred}{$\downarrow$ 3.8} & 76.3 \bf \textcolor{cssred}{$\downarrow$ 5.0} \\
                
                \hline

		\end{tabular}}
		
  \caption{\textbf{Evaluation of Closed-Set Object Detection Approaches.} We conduct evaluations using the standard COCO~\cite{lin2014microsoft} validation dataset and the proposed \dataset~dataset. We report the AP (Average Precision) and AP50 (Mean Average Precision at IoU 0.5) metrics for both classical and state-of-the-art object detectors.}
        \label{main result}
	\end{center}
\end{table*}

\section{Benchmark on \dataset}
\label{sec:benchmark}

In this section, we benchmark thoroughly both traditional deep learning-based and state-of-the-art closed-set object detectors on the proposed \dataset~dataset(Table~\ref{main result}). This includes anchor-free detectors (FCOS~\cite{tian2019fcos}, CenterNet~\cite{duan2019centernet}, YOLOX~\cite{ge2021yolox}, RTMDet~\cite{lyu2022rtmdet}, YOLOv8~\cite{Jocher_Ultralytics_YOLO_2023}, YOLOv9~\cite{wang2024yolov9}, and YOLOv10~\cite{wang2024yolov10}), anchor-based detectors (Faster R-CNN~\cite{ren2015faster}, Cascade R-CNN~\cite{cai2018cascade}, RetinaNet~\cite{lin2017focal}, YOLOv5~\cite{yolov5}, and YOLOv7~\cite{wang2023yolov7}), and transformer-based detectors (DETR~\cite{carion2020end}, Deformable-DETR~\cite{zhu2020deformable}, DAB-DETR~\cite{liu2022dabdetr}, DINO~\cite{zhang2022dino}, RT-DETR~\cite{lv2023detrs}, and Co-DETR~\cite{zong2022detrs}). Additionally, we provide the evaluation of open-set object detectors (YOLO-World~\cite{Cheng2024YOLOWorld}, GLIP~\cite{li2022grounded}, and Grounding DINO~\cite{liu2023grounding}) in Table~\ref{open-set result}. Visual results are illustrated in Figure~\ref{fig:visual}, with more visualizations included in the Appendix~\ref{sec:visualizations}.

\subsection{Implementation Details} \label{sec:setting}
We employ the MMDetection toolbox~\cite{mmdetection} for our benchmark evaluations. For detectors not implemented within the MMDetection toolbox~\cite{mmdetection}, we use their officially released models. All models are evaluated without retraining, and results are computed using the official COCO API. Links to the released models are detailed in the Appendix~\ref{sec:Links}.

\subsection{Evaluation of closed-set object detectors} \label{sec:exp_closed_set}

For closed-set object detectors, which are trained and evaluated on predefined categories, the results are presented in Table~\ref{main result}. We evaluate both CNN-based and Transformer-based detectors, with CNN-based detectors further categorized into anchor-free and anchor-based designs. From the table, several observations can be made: First, compared to the results on COCO Val~\cite{lin2014microsoft}, even though the images of \dataset~are selected using CO-DETR~\cite{zong2022detrs}, the performance of all detectors exhibits a notable decline on the proposed \dataset. Notably, RT-DETR~\cite{lv2023detrs} experiences a drop of 5.7 and 7.8 in AP and AP50, respectively, indicating a substantial number of erroneous predictions by the detector. This is a significant concern in real-world applications, highlighting the importance and necessity of our proposed dataset. Second, among CNN-based detectors, anchor-based detectors tend to exhibit a less significant performance decline compared to anchor-free detectors. This is particularly evident in two-stage methodologies, such as Cascade R-CNN~\cite{cai2018cascade}, where AP50 only registers a decrement of 2.9. Interestingly, while YOLOv5-X~\cite{yolov5} performs worse on COCO, it achieves better performance than YOLOX-X~\cite{ge2021yolox} and RTMDet-X~\cite{lyu2022rtmdet} in terms of AP50 on \dataset, suggesting that YOLOv5-X~\cite{yolov5} might be more robust when encountering background objects. Finally, among Transformer-based detectors, DINO~\cite{zhang2022dino} exhibits the fewest false positive predictions, with its AP50 on \dataset~ dropping by only 2.8. 
Surprisingly, Transformer-based detectors perform worse on \dataset~ compared to CNN counterparts. For example, RT-DETR~\cite{lv2023detrs} and YOLOv10-X~\cite{wang2024yolov10} achieve similar AP on the COCO validation set (54.3 vs. 54.4), but YOLOv10-X~\cite{wang2024yolov10} surpasses RT-DETR~\cite{lv2023detrs} by 0.8 AP on \dataset~. Overall, most detectors show inconsistent improvements between COCO validation and \dataset~.

\begin{figure*}[p]
  \centering
    \begin{subfigure}[b]{\linewidth}
         \centering
        \includegraphics[width=0.75\linewidth]{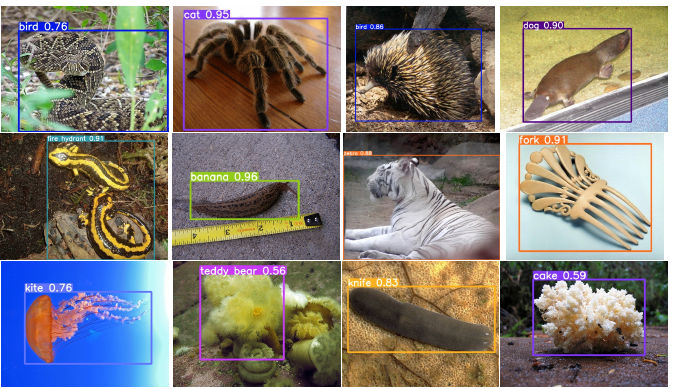}
         \caption{False positive predictions with YOLOv9-E~\cite{wang2024yolov9} on \dataset.}
         \label{fig:visual_a}
         
         \centering
        \includegraphics[width=0.75\linewidth]{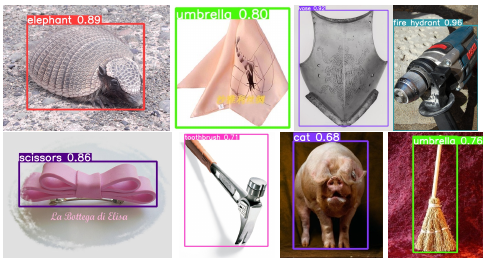}
         \caption{False positive predictions with RTMDet-X~\cite{lyu2022rtmdet} on \dataset.}
         \label{fig:visual_b}

          \centering
        \includegraphics[width=0.75\linewidth]{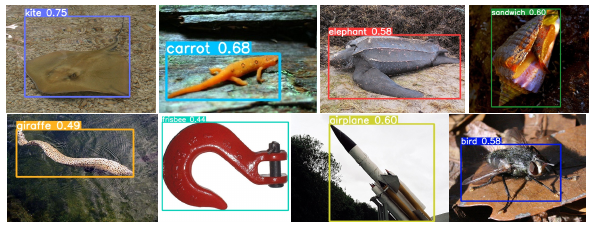}
         \caption{False positive predictions with Grounding DINO-B~\cite{liu2023grounding} on \dataset.}
         \label{fig:visual_c}
         
    \end{subfigure}
    \caption{Visualization of false positive predictions on \dataset~for different object detectors: (a) YOLOv9-E~\cite{wang2024yolov9}, (b) RTMDet-X~\cite{lyu2022rtmdet}, and (c) Grounding DINO-B~\cite{liu2023grounding} (training on COCO~\cite{lin2014microsoft} and without providing FP category as text prompt). Note that these false positive predictions have significant relatively high scores. Visual results with other object detectors are provided in the Appendix.}
    \label{fig:visual}
\end{figure*}

\subsection{Evaluation of open-set object detectors} \label{sec:exp_open_set}
For open-set object detectors that identify bounding boxes based on text prompts, we present results on COCO~\cite{lin2014microsoft} and our proposed COCO-FP in Table~\ref{open-set result}. We evaluate these detectors under two common settings: zero-shot, where the detectors have not been trained on COCO, and supervised, where they have been trained on COCO. Additionally, we assess performance by incorporating 50 novel categories (FP categories) as extra text prompts alongside the original 80 COCO categories. Given the open-set nature of these detectors, the inclusion of FP categories is expected to enhance performance, particularly in the zero-shot setting. From the table, we have the following observations: First, similar to closed-set detectors, all open-set detectors exhibit significant performance degradation, regardless of whether they are in the zero-shot setting or trained on COCO~\cite{lin2014microsoft}. Second, training or fine-tuning on the target dataset COCO~\cite{lin2014microsoft} results in substantial performance improvement, indicating that a performance gap remains between the zero-shot setting, which aims for a universal object detector, and the task-specific setting. Lastly, incorporating extra FP categories yields slight performance improvement for YOLO-World-X~\cite{Cheng2024YOLOWorld} (AP50 of COCO Val improves from 62.6 to 62.7, and AP50 of the proposed COCO-FP improves from 57.4 to 58.1) but results in performance drops for Grounding DINO-B\footnote{When incorporating FP categories, GLIP and Grounding DINO may exhibit anomalous results. See \url{https://github.com/IDEA-Research/GroundingDINO/issues/84} and \url{https://github.com/open-mmlab/mmdetection/issues/11100}}~\cite{liu2023grounding}. In general, knowing in advance the categories that might cause false positive predictions is not particularly helpful for open-set object detectors, especially if the detectors are fine-tuned on the target dataset.

\begin{table*}[t!]
    \begin{center}
		\resizebox{2\columnwidth}{!}{
			\begin{tabular}{l|l|c|c|c||c|c||c|c}
				\hline 
				\multirow{ 2}{*}{Detector}  & \multirow{ 2}{*}{Backbone} & Image&\multirow{ 2}{*}{Params}& \multirow{ 2}{*}{FP Category }  & \multicolumn{2}{c||}{COCO Val~\cite{lin2014microsoft}} & \multicolumn{2}{c}{\dataset} \\ 
                  & &Resolution  & & & AP  & AP50 & AP  & AP50 \\ \hline
				\multicolumn{9}{c}{\textbf{Zero-shot object detection}} \\
     \multirow{2}{*}{YOLO-World-X~\cite{Cheng2024YOLOWorld}} & \multirow{2}{*}{CSPDarkNet~\cite{bochkovskiy2020yolov4}} & \multirow{2}{*}{640$\times$640}& \multirow{2}{*}{136M}  &  &46.8&62.6&43.0&57.4 \\
          &&&   & \cmark &46.9&62.7&43.5&58.1\\
       \cdashline{1-9} 
       
        \multirow{2}{*}{Grounding DINO-T~\cite{liu2023grounding}}
        &\multirow{2}{*}{Swin-T~\cite{liu2021swin}}& \multirow{2}{*}{1333$\times$800}
        &\multirow{ 2}{*}{172M}  &  &48.5&64.4&44.6& 60.3\\
         &&&  & \cmark &45.4&61.4&44.5& 60.2 \\ 
        \cdashline{1-9} 
        
         \multirow{1}{*}{GLIP-L~\cite{li2022grounded}} & \multirow{1}{*}{Swin-L~\cite{liu2021swin}} & \multirow{1}{*}{1333$\times$800}& \multirow{1}{*}{430M} &  &51.3&68.2&47.7& 63.2\\
         \hline  
         
        \multicolumn{9}{c}{\textbf{Training on COCO~\cite{lin2014microsoft}}}  \\ 
        \multirow{2}{*}{YOLO-World-X~\textsuperscript{†}~\cite{Cheng2024YOLOWorld}} & \multirow{2}{*}{CSPDarkNet~\cite{bochkovskiy2020yolov4}} & \multirow{2}{*}{640$\times$640}& \multirow{2}{*}{136M} &&54.7&71.6&49.8&65.2\\
          &&&  & \cmark &54.6&71.6&49.6&65.0 \\
        \cdashline{1-9} 
        
        \multirow{2}{*}{Grounding DINO-B~\cite{liu2023grounding}}
        &\multirow{2}{*}{Swin-B~\cite{liu2021swin}}& \multirow{2}{*}{1333$\times$800}
        &\multirow{ 2}{*}{233M} &  &56.9&74.2&54.5& 71.0\\
          &&&  & \cmark &53.3&69.5&51.3& 66.9\\
        \cdashline{1-9} 
        
        \multirow{1}{*}{GLIP-L~\textsuperscript{†}~\cite{li2022grounded}} & \multirow{1}{*}{Swin-L~\cite{liu2021swin}} & \multirow{1}{*}{1333$\times$800}& \multirow{1}{*}{430M} &  &59.4&77.4&53.4& 69.6\\
         \hline 
		\end{tabular}}
		\caption{\textbf{Evaluation of Open-Set Object Detection Approaches.} We conduct evaluations using the standard COCO~\cite{lin2014microsoft} validation dataset and the proposed \dataset~dataset. We report the AP (Average Precision) and AP50 (Mean Average Precision at IoU 0.5) metrics for both zero-shot setting (without training on COCO) and training on COCO. We further conduct an evaluation assuming FP categories are known.~\textsuperscript{†} denotes finetuning on the COCO~\cite{lin2014microsoft} training set.}
        \label{open-set result}
	\end{center}
\end{table*}

\subsection{Visualization}
 \label{sec:vis}

 In Figure~\ref{fig:visual}, we present visualizations of false positive predictions made by three recent state-of-the-art object detectors on the proposed \dataset~dataset: YOLOv9-E\cite{wang2024yolov9}, RTMDet-X~\cite{lyu2022rtmdet}, and Grounding DINO-B~\cite{liu2023grounding} (training on COCO~\cite{lin2014microsoft} and without providing FP categories). YOLOv9-E~\cite{wang2024yolov9} and RTMDet-X~\cite{lyu2022rtmdet} are closed-set detectors, while Grounding DINO-B~\cite{liu2023grounding} is an open-set detector. Additional visual results with other detectors are provided in the appendix. As shown in Figure~\ref{fig:visual_a}, YOLOv9-E~\cite{wang2024yolov9} exhibits incorrect predictions across many distinct categories, often with high confidence scores. A similar pattern is observed with RTMDet-X~\cite{lyu2022rtmdet} in Figure~\ref{fig:visual_b}. Although Grounding DINO-B~\cite{liu2023grounding} is expected to recognize unknown categories, it also produces false positive predictions on the proposed COCO-FP dataset (Figure~\ref{fig:visual_c}). Notably, all presented images are relatively simple, featuring a main object and straightforward context. In real-world applications, samples are more complex, posing additional challenges for object detectors in managing false positive predictions. Our COCO-FP dataset complements the standard COCO~\cite{lin2014microsoft} benchmark by providing a quantitative assessment of performance when encountering non-target visual clutters.

\section{Discussion}
\label{sec:discussion}

\begin{figure}[!t]
  \centering
  \includegraphics[width=1\linewidth]{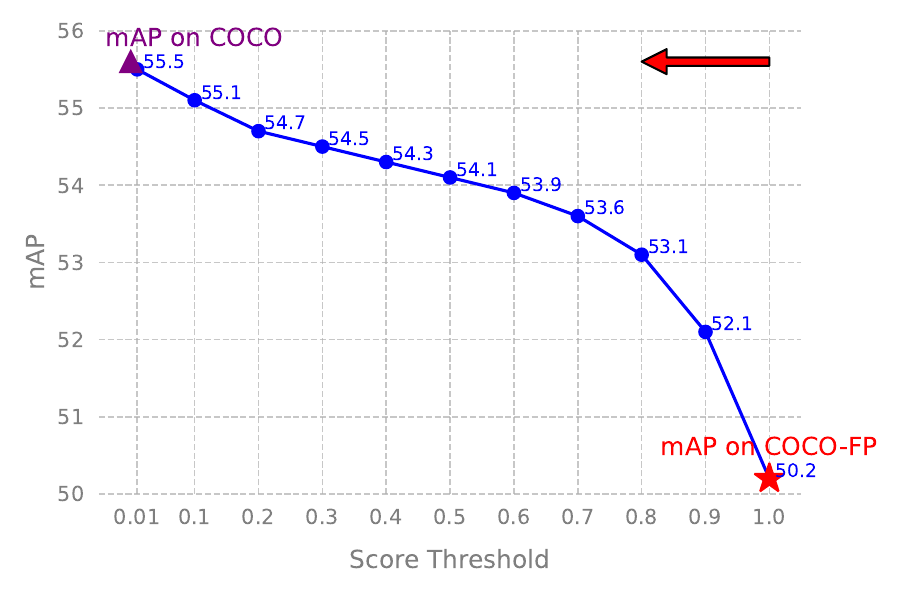}
  \caption{\textbf{The impact of false positive bounding boxes scores on mAP.} The horizontal axis represents the maximum score threshold applied to bounding boxes produced by YOLOv9-E~\cite{wang2024yolov9} from the 3,772 images, while the vertical axis shows the corresponding mAP on \dataset~dataset. This indicates that the detector produced a substantial number of high-scoring false positives, a critical issue for real-world applications.}
  \label{fig:fixerror}
\end{figure}

Intuitively, one might expect object detectors to produce no bounding box predictions on the 3,772 images in the \dataset~dataset, as these images contain no annotated objects. However, this expectation does not fully align with the calculation of mAP. Not all false positives negatively impact mAP, as the COCO dataset~\cite{lin2014microsoft} smooths the precision-recall curve by selecting the maximum precision for recall values greater than the current recall. Consequently, only false positives bounding boxes with high scores—those exceeding some true positive scores—affect mAP. Thus, detectors may still generate bounding boxes in non-target clutter regions, as long as the associated scores are sufficiently low.

Figure~\ref{fig:fixerror} illustrates our evaluation of YOLOv9-E~\cite{wang2024yolov9} on \dataset~to assess the impact of false positive scores on mAP. The horizontal axis represents the maximum score threshold applied to bounding boxes from the 3,772 images, while the vertical axis shows the corresponding mAP. Setting the score threshold to 0.3 increased YOLOv9-E's mAP on \dataset~from 50.2 to 54.5, significantly narrowing the gap with its mAP on COCO Val~\cite{lin2014microsoft}. This indicates that the detector produced a substantial number of high-scoring false positives, a critical issue for real-world applications. While score thresholds can filter out low-confidence false positives, addressing high-confidence false positives remains a significant challenge.

\section{Conclusion}
\label{sec:con}

In this paper, we introduce \dataset, a curated evaluation dataset from ImageNet-1K aimed at highlighting false positive predictions in object detection models. Unlike previous datasets, \dataset~is specifically designed to evaluate background false positives, which continue to pose significant challenges in practical scenarios. Our benchmark results reveal that state-of-the-art object detectors struggle with background-induced false positives, as indicated by a noticeable decline in performance when transitioning from the COCO dataset to the proposed \dataset. This underlines the importance of developing detection models that are better equipped to handle false alarms, particularly in high-stakes applications like fire and smoke detection, where precision is critical. In conclusion, addressing background false positives, especially those from non-target clutter, is crucial to enhancing object detectors' real-world performance. \dataset~provides a robust foundation for evaluating and improving these models.

{\small
\bibliographystyle{ieee_fullname}
\bibliography{egbib}
}

\onecolumn
\section*{Appendix}
\appendix

\section{Distribution of 50 additional categories of images collected from ImageNet}
\label{sec:distribution}

\begin{figure}[htbp!]
  \centering
  \includegraphics[width=0.9\linewidth]{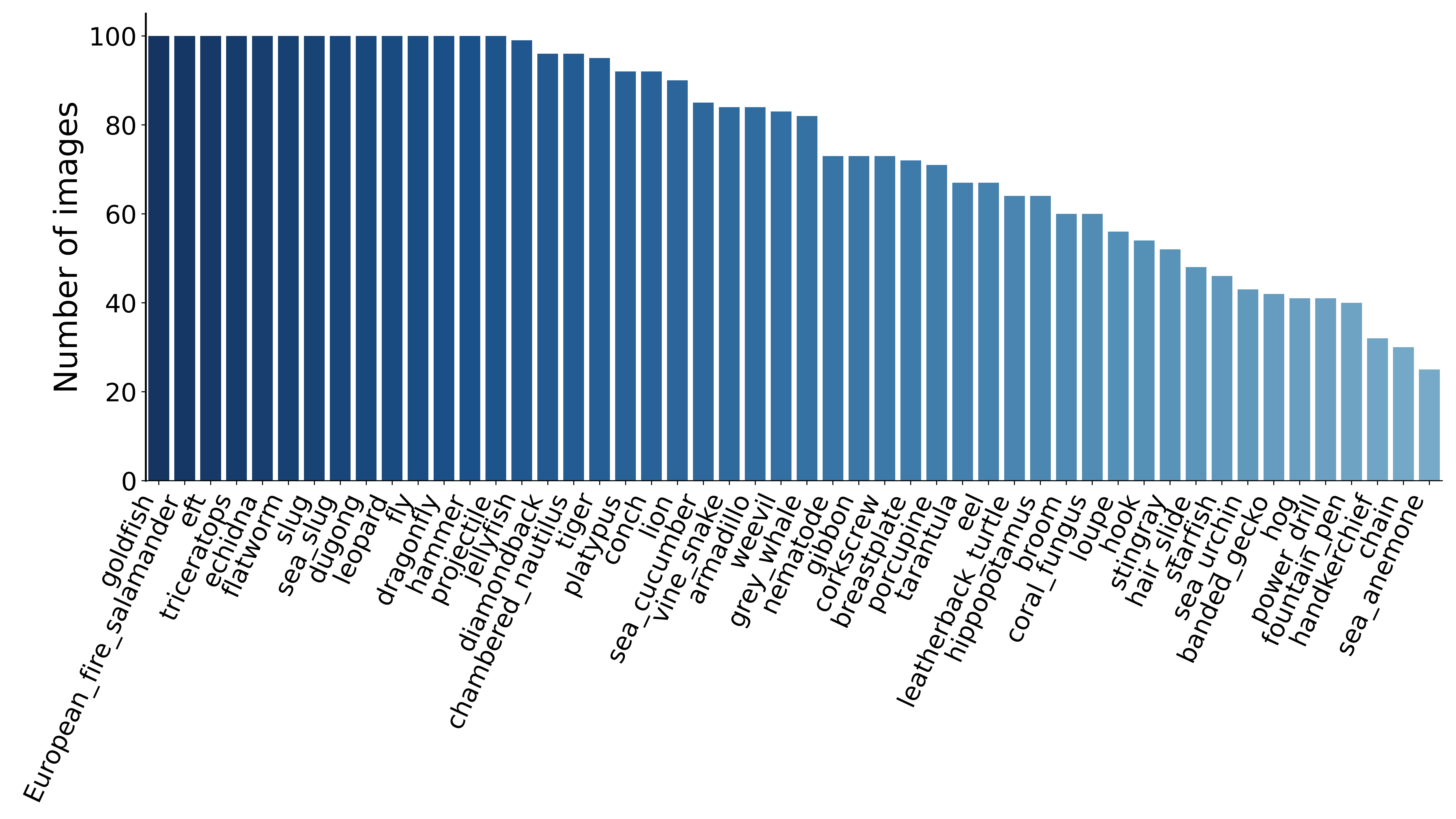}
  \caption{\textbf{The distribution of categories}. To ensure dataset diversity, we retain at most the first 100 images with the highest confidence scores in each category.}
  \label{fig:distribution}
\end{figure}

We finally collected total 3,772 images in 50 categories from ImageNet~\cite{deng2009imagenet}. In Figure \ref{fig:distribution}, 
we show the number of images in each category.

\section{Links to the officially released models used in the benchmark}
\label{sec:Links}
The official implementations we used in benchmark are as follows:

MMDetection toolbox~\cite{mmdetection}: \url{https://github.com/open-mmlab/mmdetection}  

YOLOv5~\cite{yolov5}: \url{https://github.com/ultralytics/yolov5}  

YOLOv7~\cite{wang2023yolov7}: \url{https://github.com/WongKinYiu/yolov7}  

YOLOv8~\cite{Jocher_Ultralytics_YOLO_2023}: \url{https://github.com/ultralytics/ultralytics}  

YOLOv9~\cite{wang2024yolov9}: \url{https://github.com/WongKinYiu/yolov9}  

YOLOv10~\cite{wang2024yolov10}: \url{https://github.com/THU-MIG/yolov10}  

RT-DETR~\cite{lv2023detrs}: \url{https://github.com/lyuwenyu/RT-DETR}  

YOLO-World~\cite{Cheng2024YOLOWorld}: \url{https://github.com/AILab-CVC/YOLO-World} 

Grounding DINO~\cite{liu2023grounding}: \url{https://github.com/IDEA-Research/GroundingDINO}

\section{Visualization of false positive predictions on COCO-FP}
\label{sec:visualizations}
\begin{figure}[H]
  \centering
    \begin{subfigure}[b]{0.9\linewidth}
         \centering
        \includegraphics[width=\linewidth]{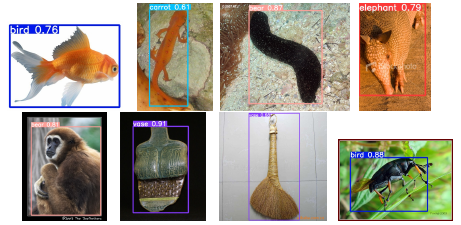}
         \caption{False positive predictions with YOLOv5-X~\cite{yolov5} on \dataset.}
         \label{fig:app_visual_a}
         
         \centering
        \includegraphics[width=0.9\linewidth]{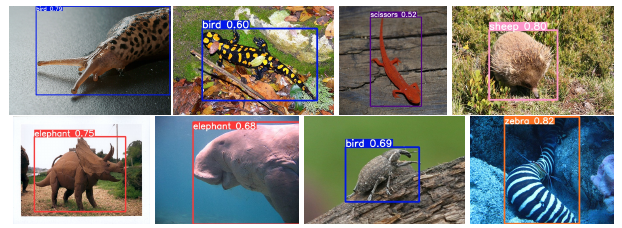}
         \caption{False positive predictions with YOLO-World-X~\cite{Cheng2024YOLOWorld} on \dataset.}
         \label{fig:app_visual_b}

          \centering
        \includegraphics[width=0.9\linewidth]{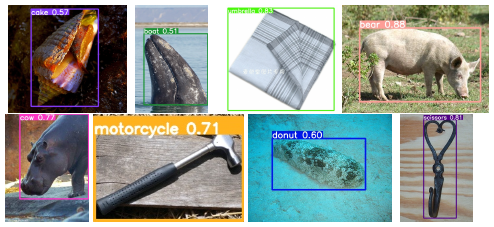}
         \caption{False positive predictions with Deformable DETR~\cite{zhu2020deformable} on \dataset.}
         \label{fig:app_visual_c}
         
    \end{subfigure}
    \caption{Visualization of false positive predictions on \dataset~for different object detectors: (a) YOLOv5-X~\cite{yolov5}, (b) YOLO-World-X~\cite{Cheng2024YOLOWorld}, and (c) Deformable DETR~\cite{zhu2020deformable}.}
    \label{fig:app_visual}
\end{figure}

\begin{figure}[H]
  \centering
  \includegraphics[width=0.9\linewidth]{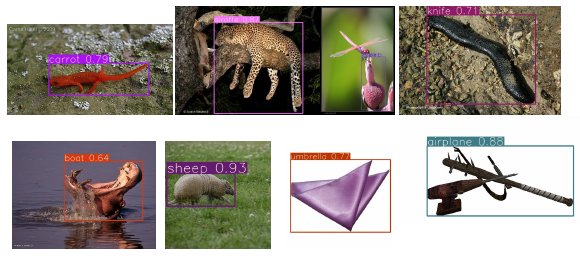}
  \caption{Visualization of false positive predictions on COCO-FP for YOLOv7-X~\cite{wang2023yolov7}.}
  \label{fig:app_visual_v7}
\end{figure}

In Figure~\ref{fig:app_visual} and Figure~\ref{fig:app_visual_v7}, we present visualizations of false positive predictions made by different object detectors on the proposed COCO-FP dataset.

\end{document}